\documentclass[11pt]{article}

\usepackage[dvipsnames]{xcolor}
\usepackage{graphicx}
\usepackage{authblk}
\newcommand{\myfootnote}[1]{
\renewcommand{\thefootnote}{}
\footnotetext{\hspace{-16.5pt}\footnotesize#1}
\renewcommand{\thefootnote}{\arabic{footnote}}}

\usepackage[utf8]{inputenc} 
\usepackage[T1]{fontenc}    
\usepackage{lmodern}
\usepackage[hidelinks]{hyperref}       
\usepackage{url}            
\usepackage{booktabs}       
\usepackage{amsfonts}       
\usepackage{nicefrac}       
\usepackage{microtype}      
\usepackage{yhmath}         
\usepackage{geometry}

\usepackage{multirow}
\usepackage{hhline}
\usepackage{color}
\usepackage{epsfig}
\usepackage{subfig}
\usepackage{graphicx}
\usepackage{amsmath}
\usepackage{amssymb}
\usepackage{amsthm}
\usepackage{bm}
\usepackage{algorithm}
\usepackage{algorithmicx}
\usepackage{algpseudocode}
\usepackage{comment}
\usepackage[numbers,square,comma]{natbib}
\usepackage{epstopdf}
\usepackage{physics}
\usepackage{setspace}
\graphicspath{{figures/}{figures/IRCUR_restaurant/}{figures/IRCUR_mall/}}

\newtheorem{theorem}{Theorem}

\newtheorem{assumption}[theorem]{Assumption}

\def\R{\mathbb{R}}

\def\BL{\bm{L}}

\def\BD{\bm{D}}
\def\BS{\bm{S}}
\def\BC{\bm{C}}
\def\BU{\bm{U}}
\def\BR{\bm{R}}
\def\BV{\bm{V}}
\def\BW{\bm{W}}
\def\BSigma{\bm{\Sigma}}
\def\BX{\bm{X}}

\def\BW{\bm{W}}
\def\cI{\mathcal{I}}
\def\cJ{\mathcal{J}}
\def\cT{\mathcal{T}}

\def\cH{\mathcal{H}}
\def\cO{\mathcal{O}}

\def\algoname{IRCUR}

\def\rank{\mathrm{rank}}
\def\supp{\mathrm{supp}}

\DeclareMathOperator*{\minimize}{\mathrm{minimize}}
\DeclareMathOperator*{\subject}{\mathrm{subject~to~}}

\begin{document}


\title{Rapid Robust Principal Component Analysis: \\CUR Accelerated Inexact Low Rank Estimation}

\author[1]{HanQin Cai}
\author[2]{Keaton Hamm}
\author[1]{Longxiu Huang}
\author[3]{Jiaqi Li}
\author[3]{Tao Wang}
\affil[1]{Department of Mathematics, \protect\\ University of California, Los Angeles,\protect\\ Los Angeles, CA, USA.\vspace{.15cm}}
\affil[2]{ Department of Mathematics,\protect\\ University of Texas at Arlington, \protect\\Arlington, TX, USA.\vspace{.15cm}}
\affil[3]{The  
		School of Data and Computer Science,\protect\\
		Sun Yat-sen University\protect\\ Guangzhou, Guangdong, China.\vspace{.15cm}}
\myfootnote{ Email addresses: hqcai@math.ucla.edu (H.Q. Cai), keaton.hamm@uta.edu (K. Hamm), huangl3@math.ucla.edu (L.X. Huang), lijq63@mail2.sysu.edu.cn(J. Li), wangtao29@mail.sysu.edu.cn (T. Wang).}

\maketitle

\begin{abstract}
Robust principal component  analysis (RPCA) is a widely used tool for dimension reduction. In this work, we propose a novel non-convex algorithm, coined  Iterated Robust CUR (\algoname), for solving RPCA problems, which dramatically improves the computational efficiency in comparison with the existing algorithms. \algoname~achieves this acceleration by employing CUR decomposition when updating the low rank component, which allows us to obtain an accurate low rank approximation via only three small submatrices. Consequently, \algoname~is able to process only the small submatrices and avoid the expensive computing on full matrix through the entire algorithm.
Numerical experiments establish the computational advantage of \algoname~over the state-of-art algorithms on both synthetic and real-world datasets.

\end{abstract}

\providecommand{\keywords}[1]
{
  \small	
  \textbf{\textit{Keywords---}}#1
}

\keywords{RPCA, principal component analysis,  CUR decomposition, low-rank modeling, outlier removal}

\section{Introduction}

Principal component analysis (PCA) is one of the fundamental tools for dimension reduction. A well-known drawback of the standard PCA approach, viz., singular value decomposition (SVD), is its high sensitivity to outliers. Robust PCA (RPCA) is designed to overcome this shortcoming and enhance the robustness of PCA to potentially corrupted data. RPCA has received a lot of attention in recent years and appears in a wide range of applications, e.g., image alignment and rectification \citep{peng2012rasl,song2015image}, face recognition \citep{luan2014extracting,wright2008robust}, feature identification \cite{hu2019dstpca,liu2014rpca}, sparse graph clustering \citep{chen2012clustering}, NMR spectroscopy signal recovery \citep{cai2019fast}\footnote{Appears later as \citep{cai2021accelerated}},  and video background subtraction \citep{jang2016primary,moore2019panoramic}.

We consider the following problem setting for RPCA: given a sparsely corrupted observational data matrix $\BD\in\R^{n_1\times n_2}$, which is the sum of the underlying low rank matrix $\BL$ and sparse outlier matrix $\BS$, our goal is to recover $\BL$ and $\BS$ simultaneously from $\BD=\BL+\BS$. Intuitively, RPCA can be modeled as a non-convex optimization problem:
\begin{equation} \label{eq:non-convex model}
    \begin{split}
    &\minimize_{\BL',\BS'} \|\BD-\BL'-\BS'\|_F \cr 
    &\subject \rank(\BL')\leq r \quad \textnormal{and} \quad \|\BS'\|_0\leq \alpha n^2
    \end{split}
\end{equation}
where $r$ is the rank of the underlying low rank matrix, the symbol $\|\cdot\|_0$ denotes the $\ell_0$-norm, and $\alpha$ is the sparsity level of the underlying  sparse outlier matrix; for ease of notation, we use $n_1=n_2=:n$ through the paper, but emphasize that all results can be easily extended to non-square matrices. 

One can see that the RPCA model should handle outliers better than the standard PCA since outliers can be extracted as the sparse matrix $\BS$; however, the solution of \eqref{eq:non-convex model} may not be unique if the low rank component is also sparse, or vice versa \citep{candes2011robust}. To ensure the uniqueness of the solution, the following assumptions are commonly made for RPCA:

\begin{assumption}[$\mu$-incoherence of $\BL$] \label{asm:incoh}
Let $\BL=\BW\BSigma\BV^T$ be the compact SVD of $\BL$. There exists a constant $\mu$ such that
\begin{equation*}
    \|\BW\|_{2,\infty}\leq\sqrt{\frac{\mu r}{n}} \qquad \textnormal{and} \qquad
    \|\BV\|_{2,\infty}\leq\sqrt{\frac{\mu r}{n}}.
\end{equation*}
\end{assumption}

\begin{assumption}[$\alpha$-sparsity of $\BS$]  \label{asm:sparse}
$\BS$ has no more than $\alpha n$ non-zero entries in each of its rows and columns.
\end{assumption}

Essentially, Assumption~\ref{asm:incoh} ensures the low rank component is not too sparse and Assumption~\ref{asm:sparse} ensures the sparse component is not locally dense. Note that some papers use random support assumptions for $\BS$ instead of Assumption~\ref{asm:sparse}; for example, Assumption~\ref{asm:sparse} will be satisfied when the support of outliers is drawn from some commonly used stochastic process \citep[Theorem~1.2]{cai2018accelerating}. With these two assumptions, the separation of $\BL$ and $\BS$, viz., RPCA, becomes a well-posed problem.

\subsection{Prior Art and Contribution}
RPCA was raised and popularized by earlier works 
\citep{candes2011robust,xu2010robust,chandrasekaran2011rank}, wherein convex relaxed formulas for RPCA were proposed and studied. Unfortunately, these earlier approaches only achieved sublinear convergence and thus are computationally intensive \citep{ma2018efficient}. 
Later, a number of non-convex approaches were studied to solve \eqref{eq:non-convex model} directly. 
In particular, \citep{netrapalli2014non} proposed an alternating projections based non-convex algorithm and an accelerated version was studied in \citep{cai2019accelerated} later. Also, a gradient descent based method was proposed in \citep{yi2016fast}, which was recently modified for accelerating with ill-conditioned problems \citep{tong2020accelerating}. All of the aforementioned non-convex methods offered linear convergence with a complexity of at least $\mathcal{O}(rn^2)$ flops per iteration. 

In this paper, we propose a novel rapid non-convex algorithm for directly solving the non-convex RPCA model \eqref{eq:non-convex model}. The key is to integrate the CUR decomposition \citep{mahoney2009cur} with the iterative RPCA framework and replace the SVD, which dramatically reduces the complexity to $\cO(r^2n\log^2(n))$ flops per iteration.

\section{Background of CUR Decomposition} \label{sec:background}
Consider a noiseless low rank data matrix.
The standard dimension reduction tools such as PCA can achieve desired approximation and compression for data; however, the approximation may lose interpretability \citep{mahoney2009cur}. One approach for addressing this difficulty is to utilize the self-expressiveness of data, that is, a set of data is generally well-represented via linear combinations of the other data points rather than in some abstract bases, e.g., the singular vectors.

 CUR decomposition is an efficient tool to maintain the interpretability in dimension reduction. The classic CUR decomposition problem asks: given a matrix $\BL\in\mathbb{R}^{n\times n}$ with rank $r$, can we decompose it into terms involving only some of its columns and some of its rows? Particularly, if $r$ columns of $\BL$ which span the column space of $\BL$ and $r$ rows which span the row space of $\BL$ are chosen, then we can obtain $\BL$ itself from these submatrices. The answer has been known to be yes for some time.

\begin{theorem}\label{thm:noiseless CUR}
Consider row and column indices $\cI,\cJ\subseteq[n]$ with $|\cI|,|\cJ|\geq r$.  Denote submatrices $\BC=\BL_{:,\cJ}$, $\BU=\BL_{\cI,\cJ}$ and $\BR=\BL_{\cI,:}$.  If $\rank(\BU)=\rank(\BL)$, then 
$\BL = \BC \BU^\dagger \BR$, where $(\cdot)^\dagger$ denotes the Moore-Penrose pseudoinverse.
\end{theorem}

Theorem \ref{thm:noiseless CUR} is essentially folklore; the reader may consult \citep{HH2020} for a history and proof.  From the theorem statement, it  naturally implies that the success of CUR decomposition heavily relies on whether the rank of the mixing submatrix $\BU$  equals that of $\BL$. In fact, with various sampling strategies, this condition can be guaranteed with high probability if we sample an appropriate number of rows and columns. For instance, Theorem~\ref{thm:uniform sample} presents the sampling complexity for uniform sampling.  

\begin{theorem}[{\citep[Theorem~1.1]{chiu2013sublinear}}] \label{thm:uniform sample}
    Let $\BL$ satisfy Assumption~\ref{asm:incoh}, and suppose we sample $|\cI|=\cO(r \log(n))$ rows and $|\cJ|=\cO(r \log(n))$ columns uniformly with replacement.  Then $\BU = \BL_{\cI,\cJ}$  satisfies $\rank(\BU)=\rank(\BL)$ with probability at least $1-\mathcal{O}(r n^{-2})$.
\end{theorem}

Various sampling strategies have been proposed for choosing $\cI$ and $\cJ$ including column length and leverage score sampling. The complexity of computing these scores is $\mathcal{O}(n^2)$ while that for uniform scores is $\mathcal{O}(1)$. Since the order of $|\cI|,|\cJ|$ required is the same under incoherence assumptions, we focus solely on uniform sampling here (for more details, see \citep[Table 1]{HH2020S}).

Note that $\BU$ is an $\cO(r \log(n)) \times \cO(r \log(n))$ matrix under uniform sampling. By Theorem~\ref{thm:noiseless CUR}, the only computational cost is incurred by calculating the pseudo-inverse of $\BU$, which requires only $\cO(r^3\log^2(n))$ flops. In contrast, computing the SVD requires $\cO(rn^2)$ flops. This confirms the computational efficiency of CUR decomposition with larger $n$ and 
smaller $r$.

We also note that CUR decompositions have been connected with sparse optimization \cite{bien2010cur}, wherein they were shown to be distinct from the standard sparse PCA.

\section{Proposed Algorithm} \label{sec:proposed algo}

In this section, we develop a novel rapid yet robust algorithm aiming to solve the non-convex RPCA problem \eqref{eq:non-convex model} directly. Within the general alternating projections framework for RPCA \citep{netrapalli2014non,cai2019accelerated}, we propose a CUR-accelerated algorithm, dubbed Iterated Robust CUR (\algoname). As summarized in Algorithm~\ref{Algo:I-CUR}, there are two phases at the $(k+1)$--st iteration of \algoname: (Phase I) we first project $\BD-\BL_k$ to the set of sparse matrices via hard thresholding to update the estimate of $\BS$, (Phase II) then project $\BD-\BS_{k+1}$ to the set of low rank matrices via CUR decomposition to update the estimate of $\BL$.
For ease of presentation, we will discuss Phase II of Algorithm~\ref{Algo:I-CUR} first, then address Phase I.

\begin{algorithm}[t]
\caption{\textbf{I}terated \textbf{R}obust \textbf{CUR} for RPCA (\algoname)} \label{Algo:I-CUR}
\begin{algorithmic}[1]
\State \textbf{Input:} 
$\BD$: observed corrupted data matrix; 
$r$: rank; 
$\varepsilon$: target precision level; 
$\zeta_0$: initial thresholding value; 
$\gamma$: thresholding decay parameter; 
$|\cI|,|\cJ|$: sampling number of rows and columns.
\State Uniformly sample row indices $\cI$ and column indices $\cJ$.
\State $\BL_0=\bm{0}, \quad \BS_0=\bm{0}, \quad k=0$
\While{$e_k > \varepsilon$ ($e_k$ is defined as in \eqref{eq:stopping criterion}) }
    \State {\color{OliveGreen}\textit{(Optional)}} Resample indices $\cI$ and $\cJ$
    \State {\color{OliveGreen}// Phase I: updating sparse component}
    \State $\zeta_{k+1}=\gamma^{k}\zeta_0 $
    \State $[\BS_{k+1}]_{:,\cJ}=\mathcal{T}_{\zeta_{k+1}}[\BD-\BL_k]_{:,\cJ}$
    \State $[\BS_{k+1}]_{\cI,:}=\mathcal{T}_{\zeta_{k+1}}[\BD-\BL_k]_{\cI,:}$
    \State {\color{OliveGreen} // Phase II: updating low rank component}
    \State $\BC_{k+1}=[\BD-\BS_{k+1}]_{:,\cJ}$
    \State $\BU_{k+1}=\cH_r([\BD-\BS_{k+1}]_{\cI,\cJ})$ 
    \State $\BR_{k+1}=[\BD-\BS_{k+1}]_{\cI,:}$
    \State $\BL_{k+1} = \BC_{k+1}\BU_{k+1}^\dagger\BR_{k+1}$ ~~~~{\color{OliveGreen} // Do not compute this step}
    \State $k=k+1$
\EndWhile
\State \textbf{Output:} $\BC_k,\BU_k,\BR_k$: CUR decomposition of $\BL$.
\end{algorithmic}
\end{algorithm}

\subsection{Phase~II: Updating the Estimate of \texorpdfstring{$\BL$}{Lg}
} 
In prior art, a common approach for updating $\BL$ is to use the truncated SVD, which can be very costly when $n$ is large. Inspired by Theorem~\ref{thm:noiseless CUR}, we instead employ CUR decomposition as an inexact low rank approximator here.
More specifically, we let 
\begin{equation}
    \begin{split}
    \BC_{k+1}&=[\BD-\BS_{k+1}]_{:,\cJ}, \quad
    \BR_{k+1}=[\BD-\BS_{k+1}]_{\cI,:}, \\
    &~~~~\textnormal{and}\quad \BU_{k+1}=\cH_r([\BD-\BS_{k+1}]_{\cI,\cJ})\\
    \end{split}
\end{equation}
where the indices $\cI,\cJ$ are generated via uniform sampling, and $\cH_r(\cdot)$ denotes the best rank $r$ approximation (i.e., truncated SVD) to the argument. So, the updated estimate of $\BL$, i.e.,
\begin{equation}
    \BL_{k+1} = \BC_{k+1}\BU_{k+1}^\dagger\BR_{k+1},
\end{equation}
is of rank $r$. Per Theorem~\ref{thm:uniform sample}, it costs $\cO(r^3 \log^2(n))$ flops to obtain $\BU_{k+1}^\dagger$. At first glance, it appears to cost $\cO(r n^2 \log(n))$ flops to form $\BL_{k+1}$ itself; however, we actually never need to form the entire $\BL$ through \algoname, but merely the CUR components of $\BL_{k+1}$ to be saved and output. Thus, Phase~II of \algoname~costs only $\cO(r^3 \log^2(n))$ flops. Note that other rank truncation methods have been studied for CUR decomposition \citep{TroppNystrom,BeckerNystrom,HH2019}, but we have found the novel method proposed here to be the most computationally efficient.

\subsection{Phase~I: Updating the Estimate of \texorpdfstring{$\BS$}{Lg}
}
As shown in \citep{cai2019fast,netrapalli2014non,cai2019accelerated}, the projection to the set of sparse matrices can be achieved via the hard thresholding operator $\cT_\zeta$ defined as:
\begin{equation}
    [\cT_{\zeta}\BX]_{i,j} =
    \begin{cases}
    \BX_{i,j} & \quad|\BX_{i,j}| >\zeta,\\
    0  & \quad\mbox{otherwise.}
    \end{cases}
\end{equation}
At each iteration, with properly chosen thresholding value $\zeta_{k+1}$, we can obtain a sparse $\BS_{k+1}$ while keeping $\|\BS-\BS_{k+1}\|_\infty$ under control. One strategy is to pick $\zeta_{k+1}\geq\|\BL-\BL_k\|_\infty$, which implies $\supp(\BS_{k+1})\subseteq\supp(\BS)$ and $\|\BS-\BS_{k+1}\|_\infty\leq2\zeta_{k+1}$. In practice, we observe that iteratively decaying thresholding values achieves great success with proper parameter tuning.

We now turn our attention to the computational complexity of Phase~I. As discussed in Phase~II, to form the CUR components for $\BL_{k+1}$, we only need the submatrices corresponding to the indices $\cI$ and $\cJ$. Therefore, there is no need to apply thresholding on the entire matrix $\BD-\BL_k$, but only on the submatrices that we need to pass to Phase~II. That is, computing and passing $[\BS_{k+1}]_{\cI,:}$ and $[\BS_{k+1}]_{:,\cJ}$ is sufficient. Consequently, we only need to have $[\BL_k]_{\cI,:}$ and $[\BL_k]_{:,\cJ}$ for this calculation, and this is the reason why we emphasize the whole matrices should never be formed in \algoname. Recall that only the CUR components of $\BL_{k}$ have been computed and saved by the previous iteration. To update $\BS$ efficiently, we compute
\begin{align}
    [\BL_k]_{\cI,:} =  [\BC_{k}]_{\cI,:}\BU_{k}^\dagger\BR_{k}    \textnormal{~~and~~} 
    [\BL_k]_{:,\cJ} =  \BC_{k}\BU_{k}^\dagger[\BR_{k}]_{:,\cJ},
\end{align}
followed by hard thresholding. Since $|\cI|=\cO(r \log(n))$
and $|\cJ|=\cO(r \log(n))$, the computational complexity for Phase~I of \algoname~is $\cO(r^2n  \log^2(n))$.

\subsection{Overall Complexity}
Moreover, the stopping criterion is developed in terms of the related computing error of the sampled submatrices:
\begin{equation}  \label{eq:stopping criterion}
    e_k = \frac{\|[\BD-\BL_{k}-\BS_{k}]_{\cI,:}\|_F + \|[\BD-\BL_{k}-\BS_{k}]_{:,\cJ}\|_F}{\|\BD_{\cI,:}\|_F+\|\BD_{:,\cJ}\|_F}.
\end{equation}
Overall, we can just save and process the submatrices through the entire algorithm. In this regard, \algoname~enjoys both a superior computational complexity (i.e., $\cO(r^2 n\log^2(n))$ per iteration) and the memory efficiency.

Finally, at the output stage of \algoname, a CUR decomposition of the estimated $\BL$ is returned to the user, which allows for better interpretation of the low rank component. 
In the case of that the user is more interested in the traditional low rank expression (i.e., SVD), we also provide an efficient method for the conversion from CUR decomposition to SVD, which is summarized in Algorithm~\ref{Algo:cur2svd}. This conversion involves two $n\times r$ QR decompositions and a $r\times r$ SVD, which lead its complexity to $\cO(r^2n )$. Hence, this conversion between the two low rank representations does not increase the overall computational complexity.

\begin{algorithm}[t]
\caption{Efficient conversion from CUR to SVD} \label{Algo:cur2svd}
\begin{algorithmic}[1]
\State \textbf{Input:} $\BC,\BU,\BR$: CUR decomposition of the matrix.
\State $[\bm{Q}_C,\bm{R}_C]=\mathrm{qr}(\BC)$
\State $[\bm{Q}_R,\bm{R}_R]=\mathrm{qr}(\BR^T)$
\State $[\BW_U,\BSigma,\BV_U]=\mathrm{svd}(\bm{R}_C\BU^\dagger\bm{R}_R^T)$
\State $\BW=\bm{Q}_C\BW_U$  
\State $\BV=\bm{Q}_R\BV_U$
\State \textbf{Output:} $\BW,\BSigma,\BV$: SVD of the matrix.
\end{algorithmic}
\end{algorithm}

\subsection{Fixed vs. Resampled Indices} \label{sec:fixed vs resample}
It is optional to resample the indices $\cI$ and $\cJ$ in every iteration and produces two variants of \algoname: \algoname-F and \algoname-R for fixed indices and resampled indices, respectively. \algoname-F has minimal required data access and therefore less runtime in reality, but we can get stuck with bad submatrices if unlucky, although the chance is very low. On the other hand, \algoname-R uses more redundant data from different submatrices and thus can correct any one-time bad-luck situations. One can expect \algoname-R to tolerate corruptions better than \algoname-F. However, more data access may be forbidden in some application scenarios and also result a bit more computing (e.g., $\|\BD_{\cI,:}\|_F$ in \eqref{eq:stopping criterion} will be re-computed for every iteration). 

Generally speaking, one should use \algoname-F if the data access is restricted and extremely fast speed is desired; \algoname-R should be used when data access is inexpensive and best corruption tolerance is needed. 
We will further study the empirical performance of these two variants in Section~\ref{sec:numerical}.

\subsection{Parameter Tuning} \label{sec:parameter tuning}

There are a few parameters that need to be tuned in \algoname. First, for the hard thresholding operator to work properly, we need to take care of two parameters: the initial  thresholding value $\zeta_0$, and the thresholding decay parameter $\gamma$. Intuitively, $\zeta_0$ should be a positive number that helps us to filter out the irregular values immediately at initialization. An ideal choice of $\zeta_0$ is $\|\BL\|_\infty$, the maximum magnitude of the underlying low rank component, which implies $\supp(\BS_0)\subseteq\supp(\BS)$ while keeping the entrywise estimation error of $\BS$ under control. We can easily estimate $\|\BL\|_\infty$ by prior knowledge in many real-world problems. For example, in the image and video related applications, any pixels that fall out of the normal range of $[0,255]$ should be considered outliers immediately. 

On the other hand, the decay parameter $\gamma$ should be a value in $(0,1)$ that reflects the anticipated convergence rate. Generally speaking, harder problems (e.g., larger $\mu$, $r$, $\alpha$, etc.) will require larger $\gamma$ for successful recovery, while easier problems can be more quickly recovered with smaller $\gamma$ and also workable with larger $\gamma$. Using a larger value for $\gamma$ may cause slower convergence but can enhance the robustness of \algoname. Empirically, it is recommended that $\gamma\in[0.6,0.9]$.

The other parameters to be tuned are the numbers of rows and columns to be sampled. As stated in Theorem~\ref{thm:uniform sample}, it requires $|\cI|=c_I r \log(n)$ and $|\cJ|=c_J r \log(n)$ for some $c_I,c_J>0$. 
Typically, in the presence of noise, more rows and columns need to be sampled to provide robustness (e.g., \citep{aldroubi2019cur}).
In our case, larger $\alpha$ increases the minimum requirement of $c_I$ and $c_J$. We will further study the empirical relationship between $\alpha$ and the sampling constants in Section~\ref{sec:numerical}.

\section{Numerical Experiments} \label{sec:numerical}
In this section, we compare the empirical performance of two versions of \algoname, with fixed indices (\algoname-F) and with resampled indices (\algoname-R), against the state-of-the-art RPCA algorithms, AccAltProj \citep{cai2019accelerated} and GD \citep{yi2016fast}. Due to the page limit, we defer the details of experimental setting to the supplementary document.
Moreover, we provide a sample Matlab code for \algoname~at:
\begin{center}
\url{https://github.com/caesarcai/IRCUR}.
\end{center}

\subsection{Synthetic Datasets}
The experiments in this section follow the setup as in \citep{candes2011robust,netrapalli2014non,cai2019accelerated,yi2016fast}, wherein square $\BD,\BL,\BS\in\mathbb{R}^{n\times n}$ are used for demonstration. Thus, we sample same number of rows and columns in the following experiments, i.e., $|\cI|=|\cJ|=c_I r\log(n)$, but $\cI$ may not equal to $\cJ$.

\begin{figure}[t]
\centering
\subfloat{\includegraphics[width=.325\linewidth,height=.25\linewidth]{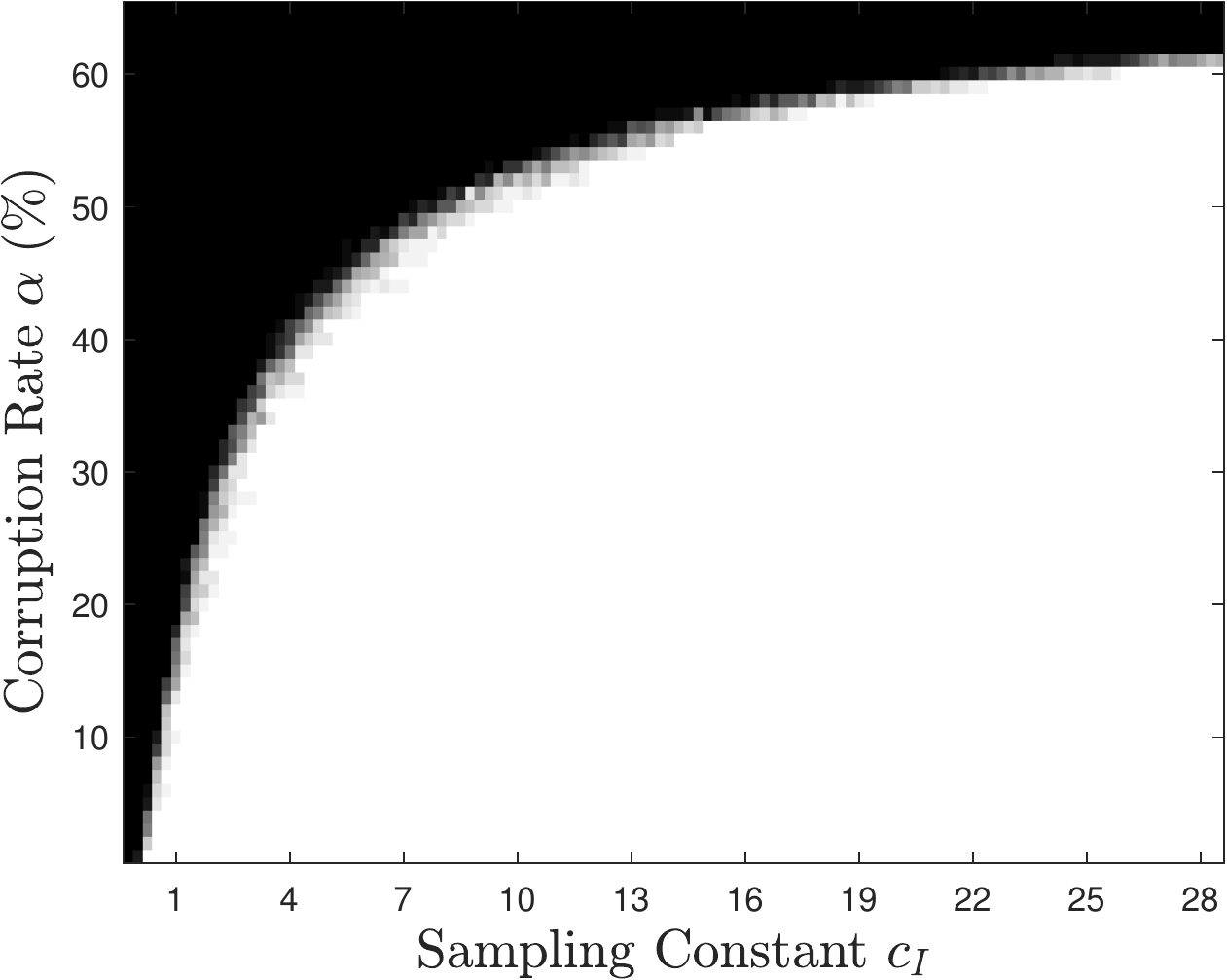}} \hfill
\subfloat{\includegraphics[width=.325\linewidth,height=.25\linewidth]{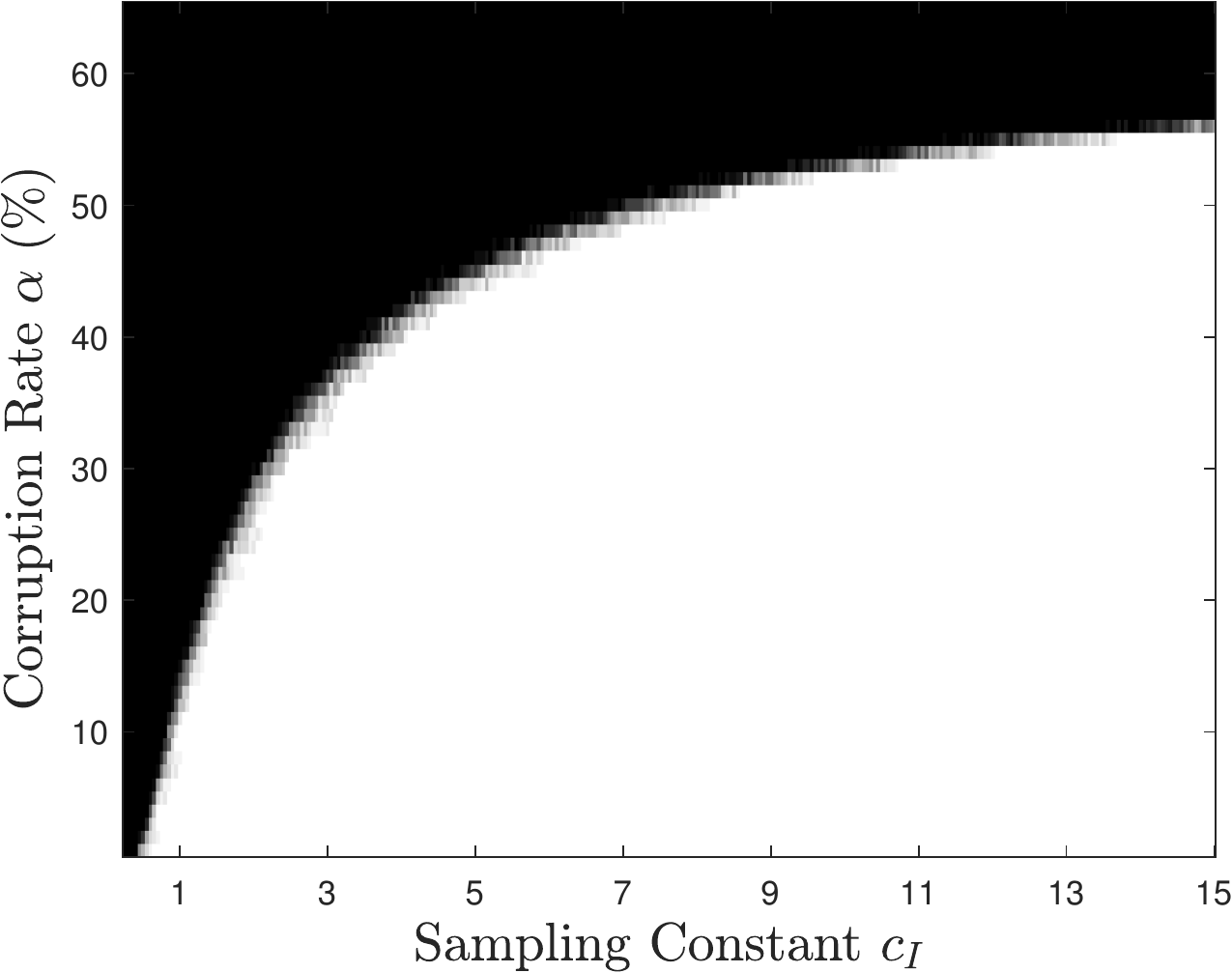}} \hfill
\subfloat{\includegraphics[width=.325\linewidth,height=.25\linewidth]{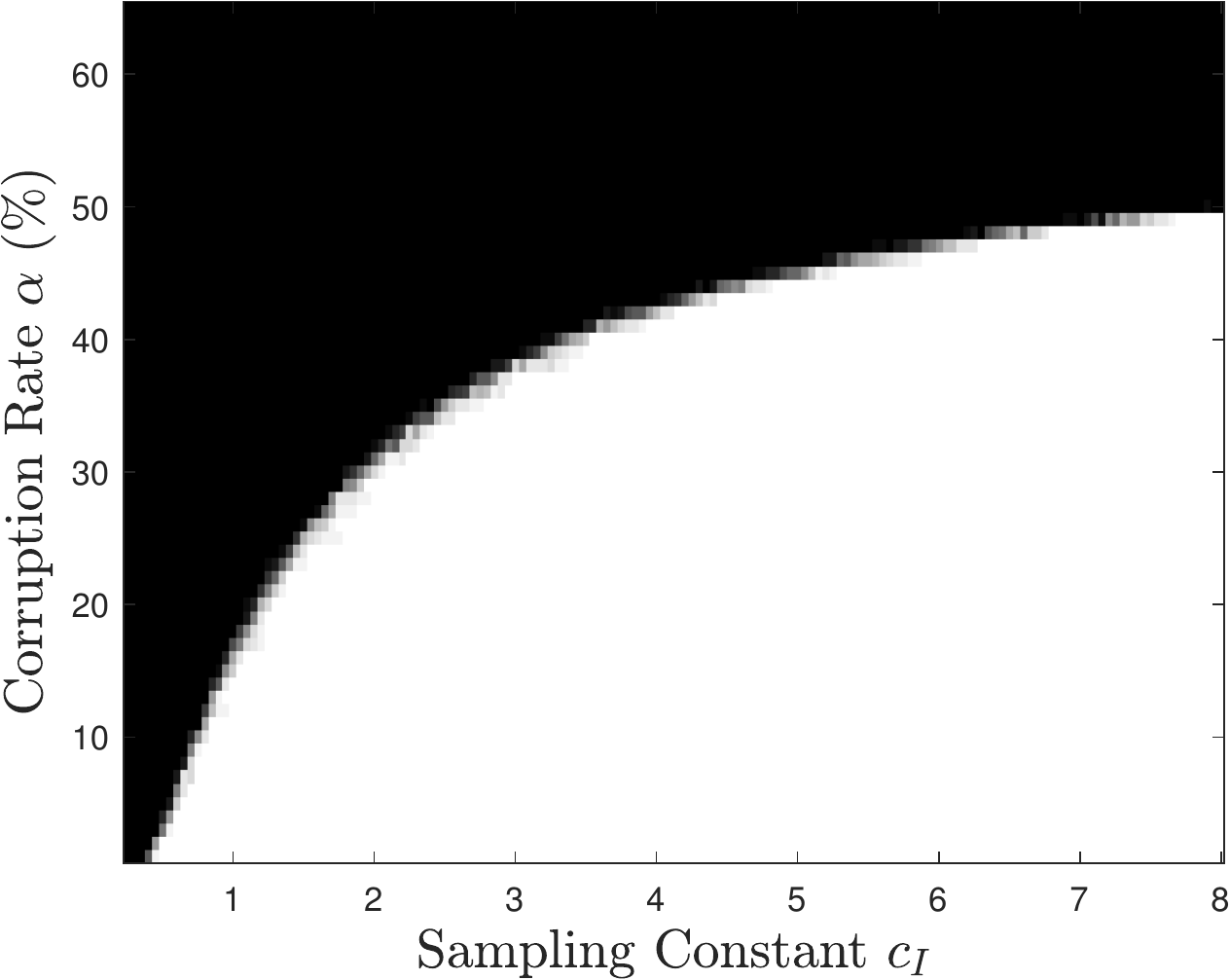}}

\subfloat{\includegraphics[width=.325\linewidth,height=.25\linewidth]{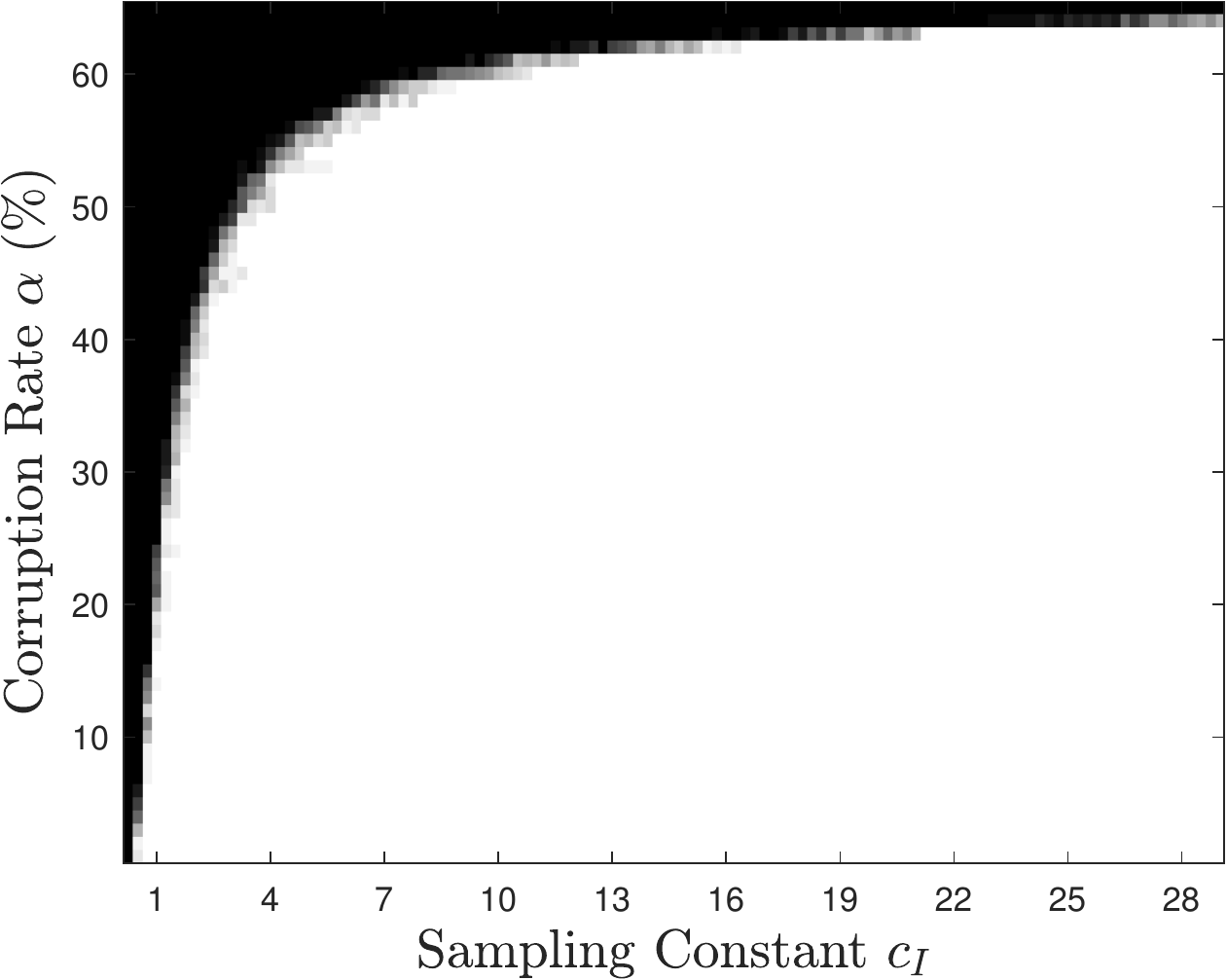}} \hfill
\subfloat{\includegraphics[width=.325\linewidth,height=.25\linewidth]{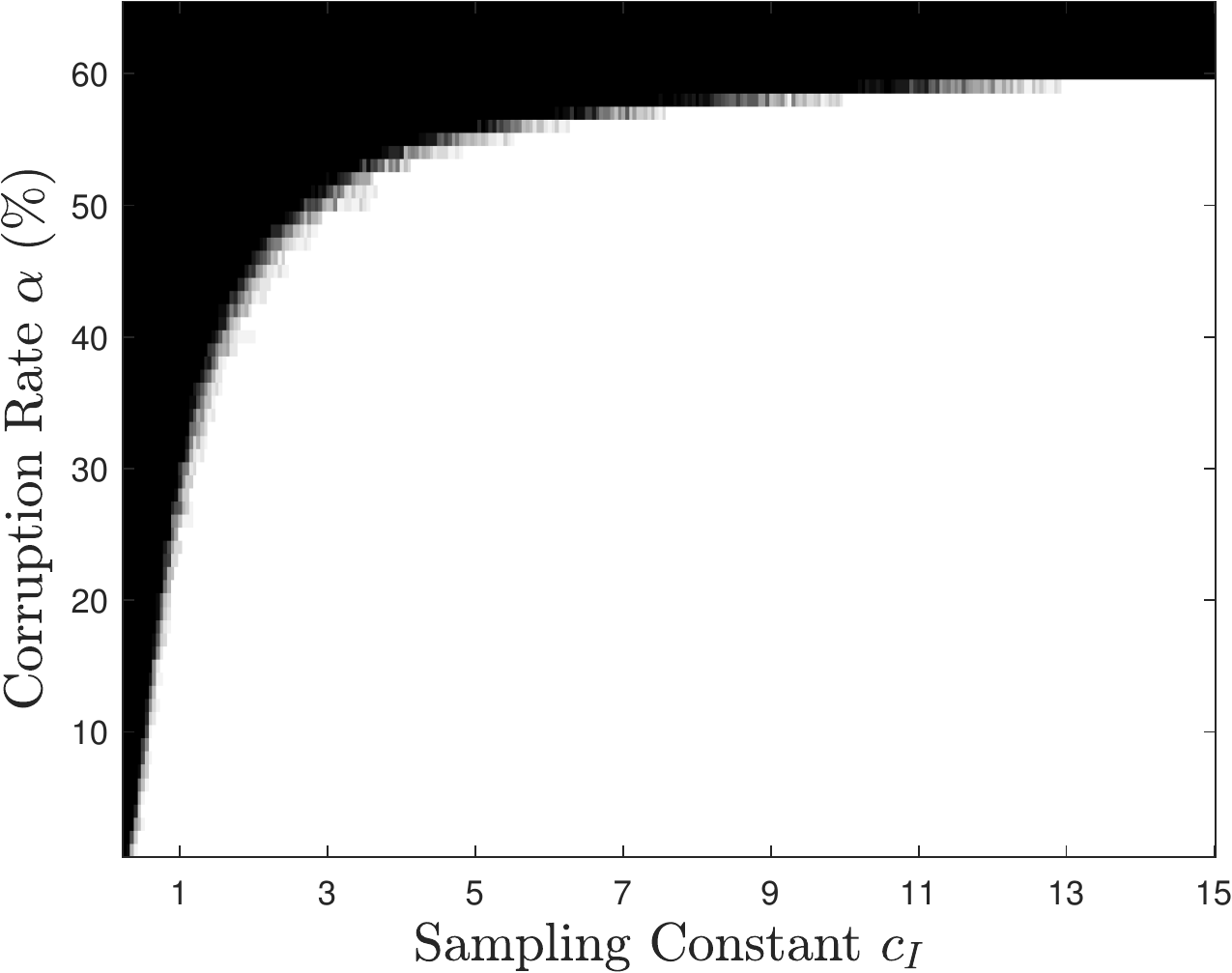}} \hfill
\subfloat{\includegraphics[width=.325\linewidth,height=.25\linewidth]{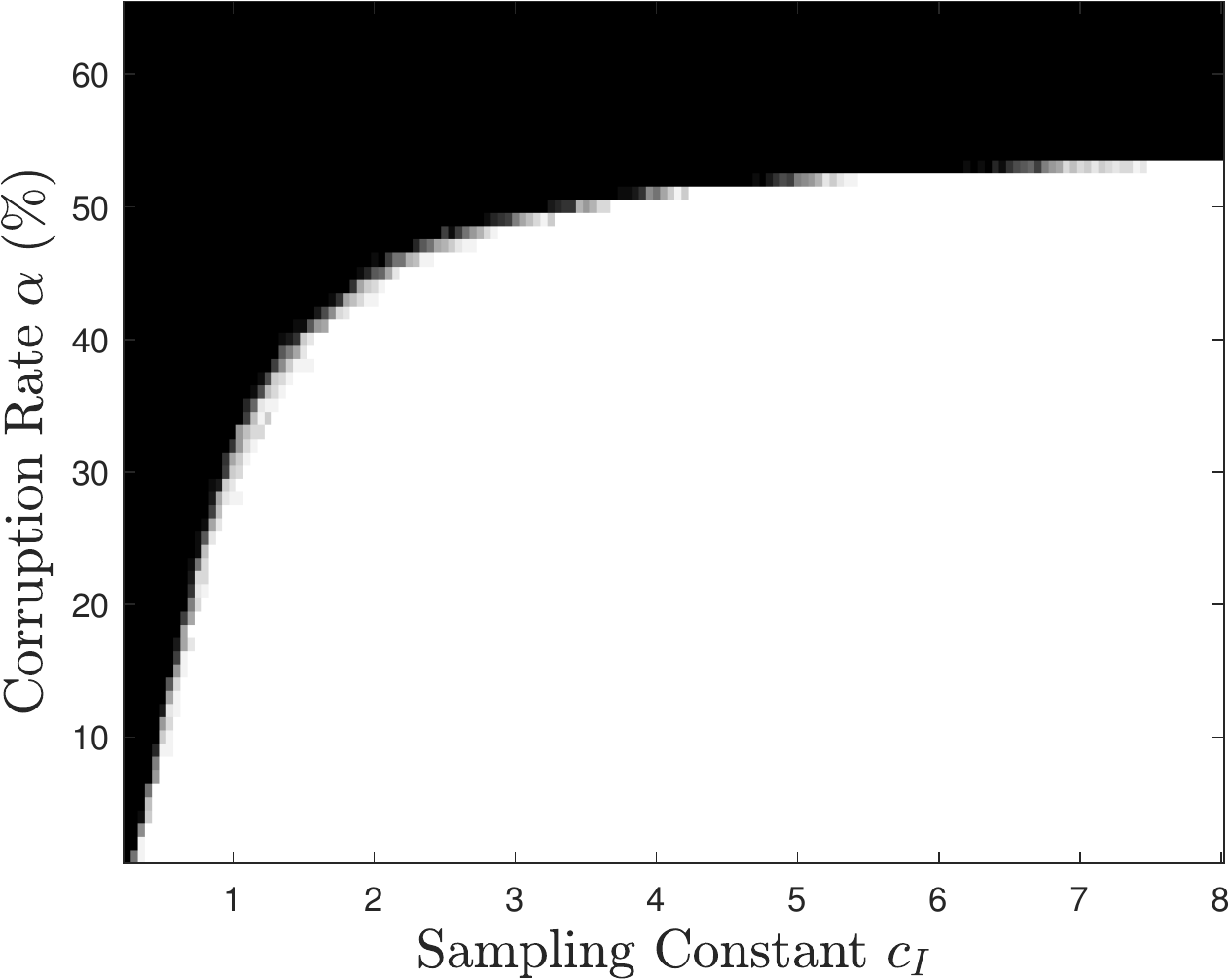}}
\caption{Empirical phase transition in sampling constant $c_I$ and corruption rate $\alpha$. A white pixel means all $50$ test cases are successfully recovered and a black pixel means all $50$ test cases fail the recovery. \textbf{Top:} \algoname-F. 
\textbf{Bottom:} \algoname-R. 
\textbf{Left: } $r=5$. \textbf{Middle: } $r=10$. \textbf{Right: } $r=20$. 
}
\label{fig:phase transition}
\end{figure}

\begin{figure}[t]
\centering
\subfloat{\includegraphics[width=.33\linewidth,height=.265\linewidth]{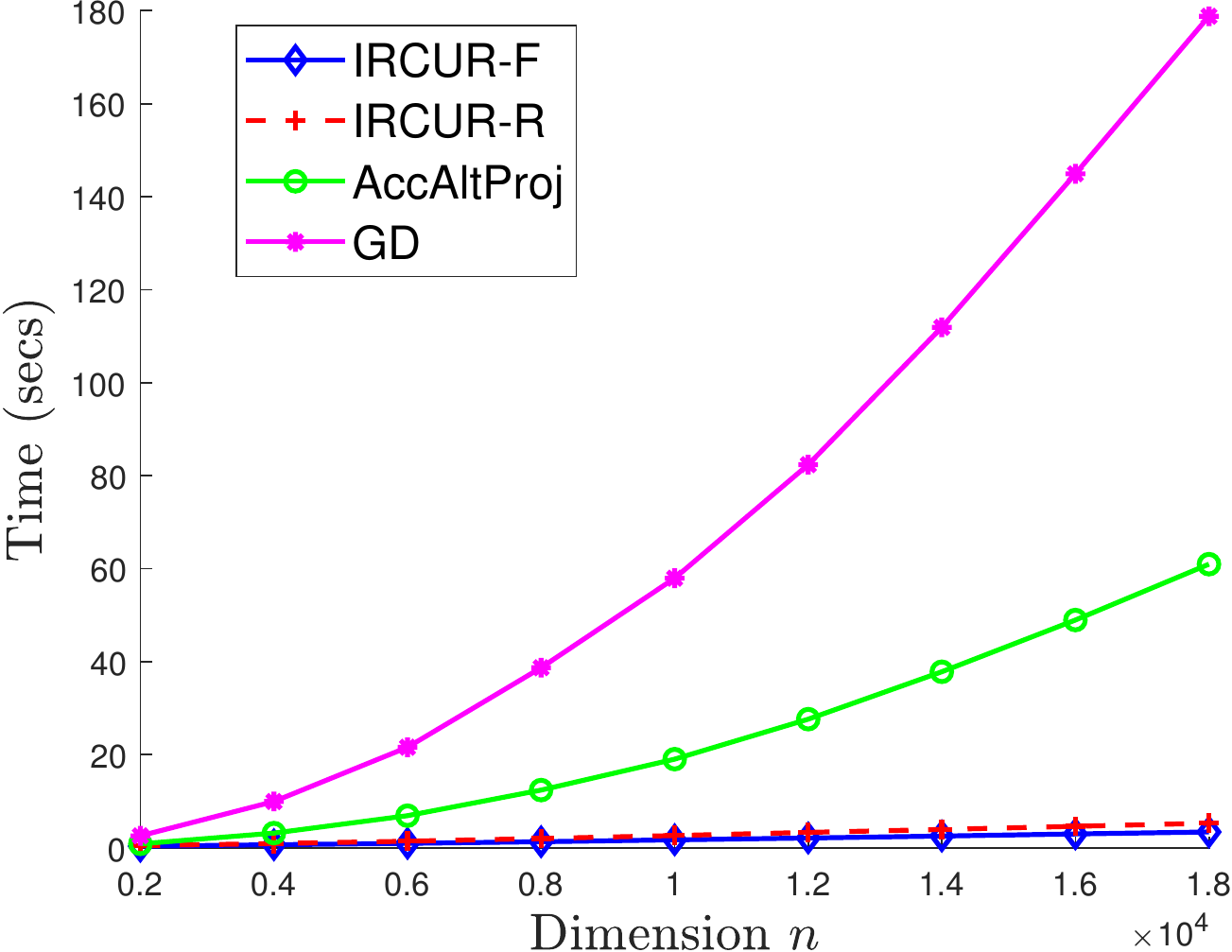}} \hfill
\subfloat{\includegraphics[width=.33\linewidth,height=.265\linewidth]{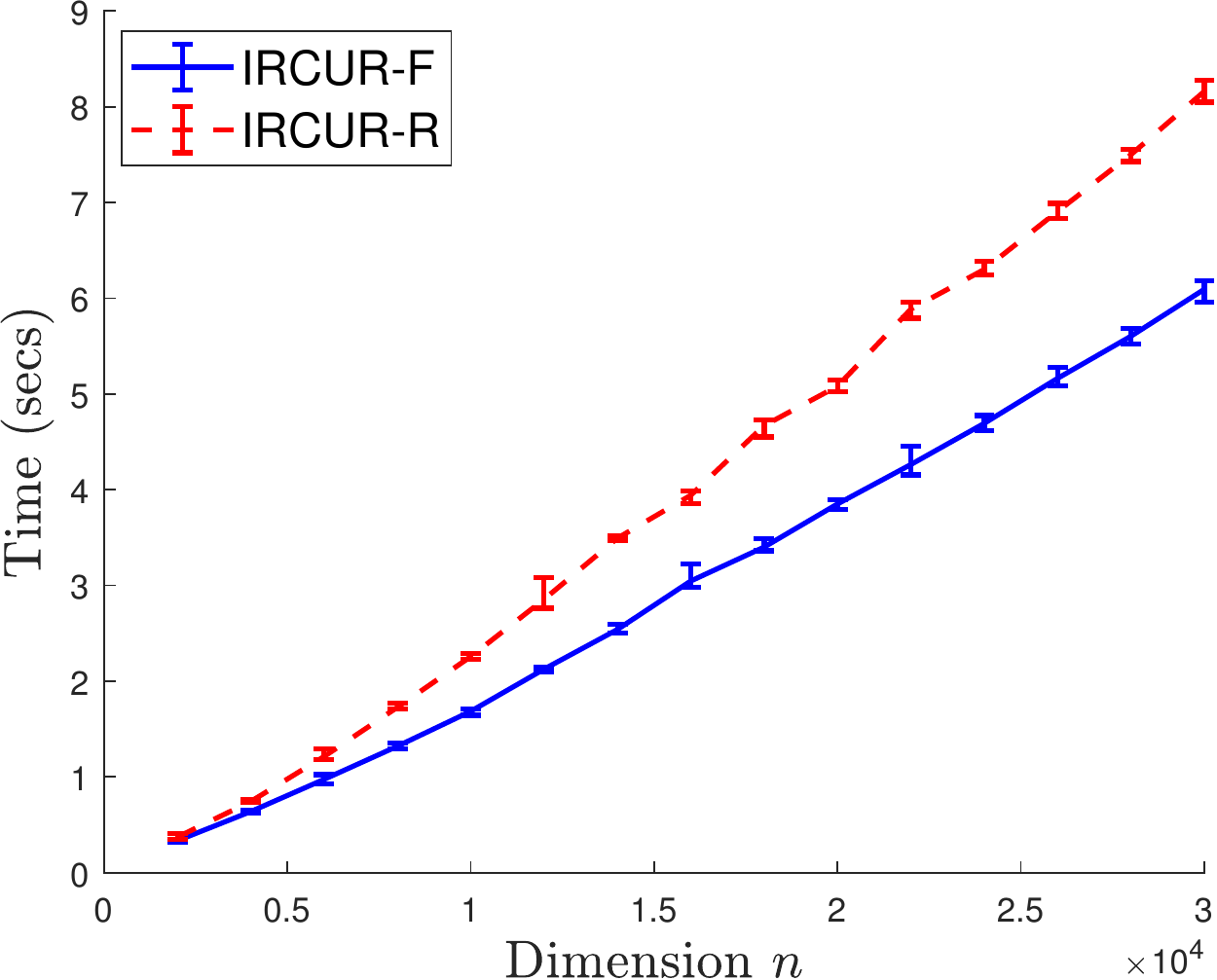}} \hfill
\subfloat{\includegraphics[width=.33\linewidth,height=.265\linewidth]{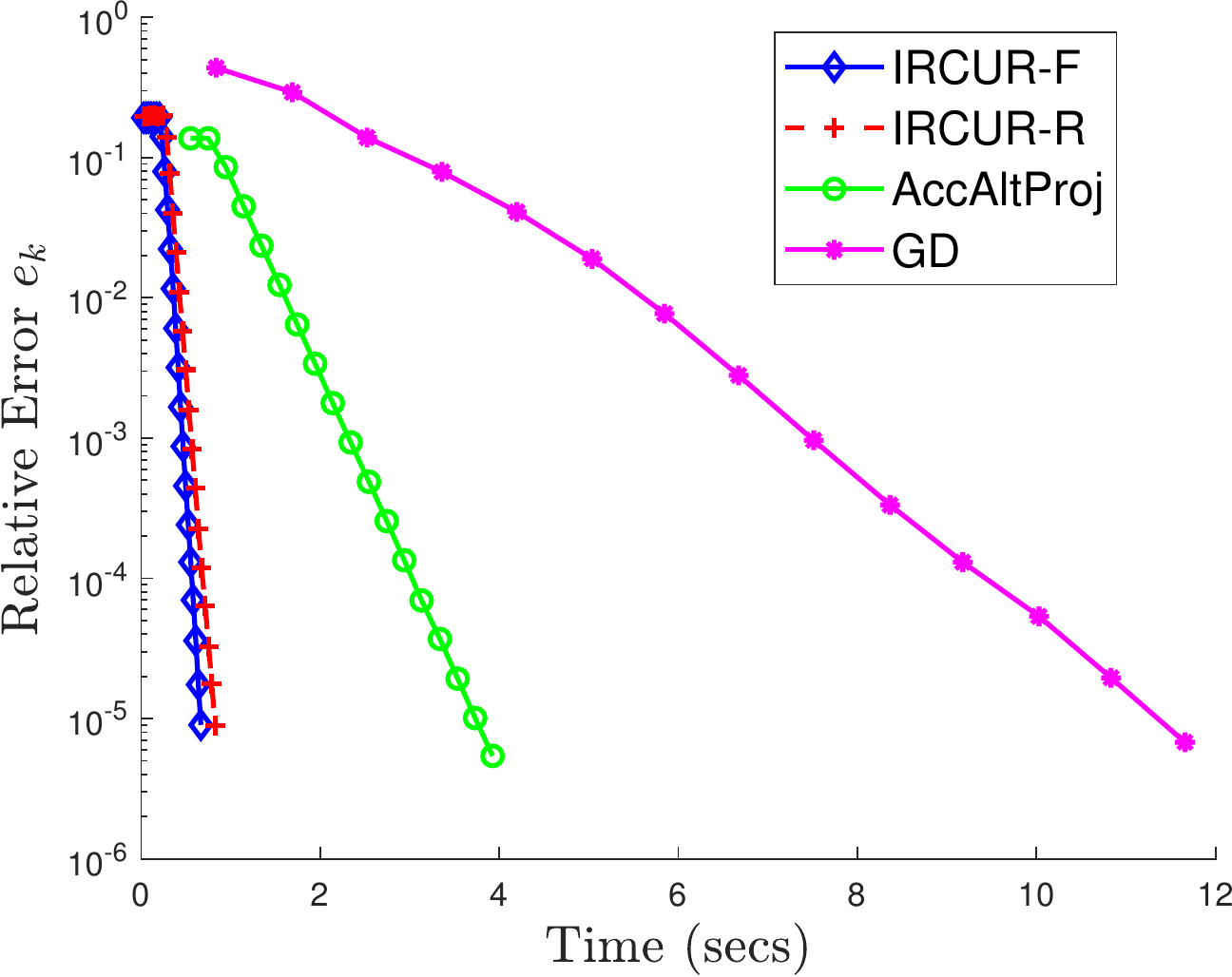}}
\caption{Runtime comparison. \textbf{Left: } dimension $n$ vs runtime: $r=5$, $\alpha=0.1$, $c_I =4$, and $n\in[2000,18000]$. \textbf{Middle: }dimension $n$ vs runtime for only \algoname: $r=5$, $\alpha=0.1$, $c_I =4$, and $n\in[2000,30000]$. \textbf{Right: } relative error $e_k$ vs runtime: $ r=5$, $\alpha=0.1$, $c_I=4$, and $n=4000$.  
}  \label{fig:speed_comparsion}
\end{figure}

\subsubsection{Empirical phase transition}
We investigate the recovery ability of \algoname~with different sampling constants $c_I$ and corruption rates $\alpha$. Taking the problem dimension $n=1000$, this experiment runs under $3$ different rank settings: $r=\{5,10,20\}$.
For each rank, we conduct $50$ random tests for every given pair of $(c_I,\alpha)$, and a recovery is deemed successful if the output of \algoname~satisfies $\|\BC_k\BU_k^\dagger \BR_k-\BL\|_F/\|\BL\|_F\leq10^{-3}$. The test results are summarized as Figure~\ref{fig:phase transition}, whereas we observe that if more rows/columns are sampled, \algoname~can handle more corruption/outliers. While the increment of sampling constant enhances the robustness rapidly when $c_I$ is small, the tolerance of corruption asymptotically reaches its limit later. This suggests that one should pick a medium value for $c_I$ to ensure robustness and efficient implementation, then $c_I\in[3,5]$ is a good balanced choice in practice. We also observe that \algoname-R produces slightly better recovery than \algoname-F as expected. 
On the other hand, the maximum tolerance of corruption gets lower as the rank increases since a higher rank results in harder problems.  

\subsubsection{Computational efficiency}
We evaluate the computational efficiency of the algorithms tested. The experimental settings and results are summarized in Figure~\ref{fig:speed_comparsion}. 
The left subfigure shows that both variants of \algoname~have substantial speed advantage against AccAltProj and GD, especially when $n$ is larger. 
The middle subfigure confirms that the computational complexity of \algoname~is indeed $\mathcal{O}(n\log^2(n))$ and \algoname-F is sightly more efficient than \algoname-R.
The right subfigure shows the linear convergence of all the algorithms tested, wherein \algoname~has the lowest runtime per iteration.

\subsection{Video Background Subtraction}

We apply the same algorithms to the task of video background subtraction. Two popular videos, \textit{shoppingmall} and  \textit{restaurant}, are used as benchmarks.
The video size information and runtime results for all test algorithms are recorded in Table~\ref{tab:video}.  We again confirm that both variants of \algoname~are substantially faster than AccAltProj and GD in this real-world benchmark.

\begin{figure}[t]
\centering
\subfloat{\includegraphics[width=.195\linewidth]{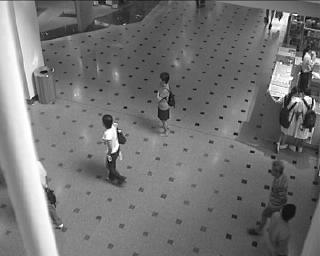}} \hfill
\subfloat{\includegraphics[width=.195\linewidth]{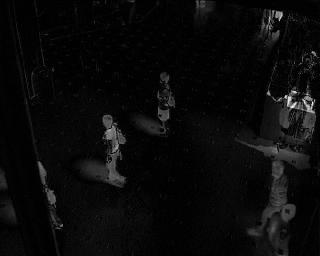}} \hfill
\subfloat{\includegraphics[width=.195\linewidth]{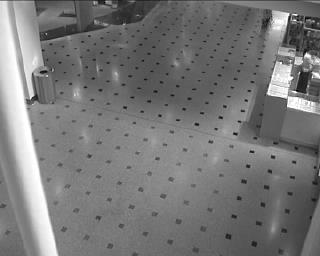}} \hfill
\subfloat{\includegraphics[width=.195\linewidth]{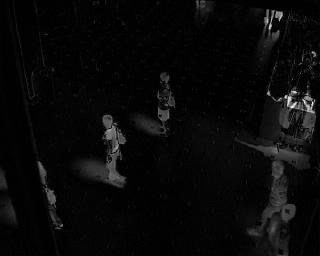}} \hfill
\subfloat{\includegraphics[width=.195\linewidth]{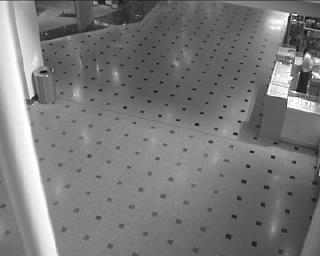}}
\vspace{-0.12in}

\subfloat{\includegraphics[width=.195\linewidth]{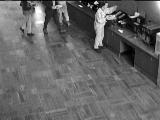}} \hfill
\subfloat{\includegraphics[width=.195\linewidth]{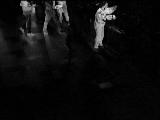}} \hfill
\subfloat{\includegraphics[width=.195\linewidth]{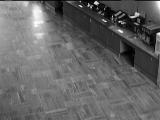}} \hfill
\subfloat{\includegraphics[width=.195\linewidth]{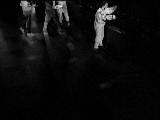}} \hfill
\subfloat{\includegraphics[width=.195\linewidth]{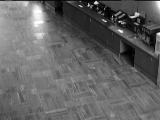}}

\caption{Video background subtraction results. The first corresponds to a frame from \textit{shoppingmall}, and the second row is from \textit{restaurant}. The first column is the original frame, the next four columns are the foreground and background outputted by \algoname-F and \algoname-R, respectively. 
}\label{FIG:video background subtraction}
\end{figure}

Moreover, all the tested algorithms achieve visually desirable results under aforementioned parameter setting. For \algoname, we present the separated background and foreground for a selected frame from each video in Figure~\ref{FIG:video background subtraction}. One can see that both variants of \algoname~enjoy crisp static backgrounds in both videos. Due to the page limit, we defer more visual results 
to the supplementary document.

\begin{table}[t]
\caption{Video size and runtime. Herein \textbf{S} represents \textit{shoppingmall} and \textbf{R} represents \textit{restaurant}.}\label{tab:video}
 \centering
 \begin{tabular}{ |c||c|c|c|c|c|c|} 
\hline
 ~             &frame & frame &   \multicolumn{4}{c|}{runtime (sec)} \cr
 \cline{4-7}

~             & size                           & number                             & \algoname-F &\algoname-R& AccAltProj    &GD             \cr

 \hhline {|=||=|=|=|=|=|=|}

\textbf{S}  &$256\times 320$              & $1000$                     &   $2.03$ &  $2.16$    & $23.04$   &$93.18$  \cr

\textbf{R}    &$120\times 160$              & $3055$                  &   $0.82$   &  $0.88$    & $15.96$&  $58.37$ \cr
\hline
\end{tabular}
\end{table}

\section{Conclusion and Future Direction} \label{sec:conclusion}

This paper presents a novel RPCA algorithm, coined \algoname, that has high computational and memory efficiency.
We use a novel CUR decomposition for rapid inexact low rank approximation, which reduces the computational complexity from typical $\cO( rn^2)$ to $\cO(r^2 n\log^2(n))$. 
The numerical simulations verify the claimed advantages of \algoname.

There are two directions for future research. First, we will investigate the theoretical convergence guarantee of \algoname~in the future. Second, it is interesting to study the recovery stability of \algoname~to additive dense noise.

\section*{Acknowledgements}

This work was supported in part by the Key-Area Research and Development Program of Guangdong Province under Grant 2020B010166001, in part by the AFOSR MURI under Grant FA9550-18-10502, in part by the ONR Grant N0001417121, in part by the ARO Grant W911NF-20-1-0076, in part by the NSF TRIPODS Grant CCF-1740858, in part by the CAREER DMS Grant 1348721, in part by the BIGDATA Grant 1740325, in part by the the Program for Guanddong Introducing Innovative and Entrepreneurial Teams under Grant 2016ZT06D211, and in part by the Cultivation Project of Supercomputing Applications under Grant 67000-18843409

\vspace{0.2in}
\bibliographystyle{unsrt}
\bibliography{ref}

\begin{thebibliography}{10}

\bibitem{peng2012rasl}
Yigang Peng, Arvind Ganesh, John Wright, Wenli Xu, and Yi~Ma.
\newblock {RASL}: Robust alignment by sparse and low-rank decomposition for
  linearly correlated images.
\newblock {\em IEEE transactions on pattern analysis and machine intelligence},
  34(11):2233--2246, 2012.

\bibitem{song2015image}
Wenjie Song, Jianke Zhu, Yang Li, and Chun Chen.
\newblock Image alignment by online robust {PCA} via stochastic gradient
  descent.
\newblock {\em IEEE Transactions on Circuits and Systems for video Technology},
  26(7):1241--1250, 2015.

\bibitem{luan2014extracting}
Xiao Luan, Bin Fang, Linghui Liu, Weibin Yang, and Jiye Qian.
\newblock Extracting sparse error of robust {PCA} for face recognition in the
  presence of varying illumination and occlusion.
\newblock {\em Pattern Recognition}, 47(2):495--508, 2014.

\bibitem{wright2008robust}
John Wright, Allen~Y Yang, Arvind Ganesh, S~Shankar Sastry, and Yi~Ma.
\newblock Robust face recognition via sparse representation.
\newblock {\em IEEE transactions on pattern analysis and machine intelligence},
  31(2):210--227, 2008.

\bibitem{hu2019dstpca}
Yue Hu, Jin-Xing Liu, Ying-Lian Gao, and Junliang Shang.
\newblock {DSTPCA}: Double-sparse constrained tensor principal component
  analysis method for feature selection.
\newblock {\em IEEE/ACM Transactions on Computational Biology and
  Bioinformatics}, 2019.

\bibitem{liu2014rpca}
Jin-Xing Liu, Yong Xu, Chun-Hou Zheng, Heng Kong, and Zhi-Hui Lai.
\newblock {RPCA}-based tumor classification using gene expression data.
\newblock {\em IEEE/ACM Transactions on Computational Biology and
  Bioinformatics}, 12(4):964--970, 2014.

\bibitem{chen2012clustering}
Yudong Chen, Sujay Sanghavi, and Huan Xu.
\newblock Clustering sparse graphs.
\newblock In {\em Advances in neural information processing systems}, pages
  2204--2212, 2012.

\bibitem{cai2019fast}
HanQin Cai, Jian-Feng Cai, Tianming Wang, and Guojian Yin.
\newblock Fast and robust spectrally sparse signal recovery: A provable
  non-convex approach via robust low-rank {Hankel} matrix reconstruction.
\newblock {\em arXiv:1910.05859}.

\bibitem{cai2021accelerated}
HanQin Cai, Jian-Feng Cai, Tianming Wang, and Guojian Yin.
\newblock Accelerated structured alternating projections for robust spectrally
  sparse signal recovery.
\newblock {\em IEEE Transactions on Signal Processing}, 69:809--821, 2021.

\bibitem{jang2016primary}
Won-Dong Jang, Chulwoo Lee, and Chang-Su Kim.
\newblock Primary object segmentation in videos via alternate convex
  optimization of foreground and background distributions.
\newblock In {\em Proceedings of the IEEE conference on computer vision and
  pattern recognition}, pages 696--704, 2016.

\bibitem{moore2019panoramic}
Brian~E Moore, Chen Gao, and Raj~Rao Nadakuditi.
\newblock Panoramic robust {PCA} for foreground--background separation on
  noisy, free-motion camera video.
\newblock {\em IEEE Transactions on Computational Imaging}, 5(2):195--211,
  2019.

\bibitem{candes2011robust}
Emmanuel~J Cand{\`e}s, Xiaodong Li, Yi~Ma, and John Wright.
\newblock Robust principal component analysis?
\newblock {\em Journal of the ACM}, 58(3):1--37, 2011.

\bibitem{cai2018accelerating}
HanQin Cai.
\newblock {\em Accelerating truncated singular-value decomposition: a fast and
  provable method for robust principal component analysis}.
\newblock PhD thesis, University of Iowa, 2018.

\bibitem{xu2010robust}
Huan Xu, Constantine Caramanis, and Sujay Sanghavi.
\newblock Robust {PCA} via outlier pursuit.
\newblock In {\em Advances in neural information processing systems}, pages
  2496--2504, 2010.

\bibitem{chandrasekaran2011rank}
Venkat Chandrasekaran, Sujay Sanghavi, Pablo~A Parrilo, and Alan~S Willsky.
\newblock Rank-sparsity incoherence for matrix decomposition.
\newblock {\em SIAM Journal on Optimization}, 21(2):572--596, 2011.

\bibitem{ma2018efficient}
Shiqian Ma and Necdet~Serhat Aybat.
\newblock Efficient optimization algorithms for robust principal component
  analysis and its variants.
\newblock {\em Proceedings of the IEEE}, 106(8):1411--1426, 2018.

\bibitem{netrapalli2014non}
Praneeth Netrapalli, UN~Niranjan, Sujay Sanghavi, Animashree Anandkumar, and
  Prateek Jain.
\newblock Non-convex robust {PCA}.
\newblock In {\em Advances in Neural Information Processing Systems}, pages
  1107--1115, 2014.

\bibitem{cai2019accelerated}
HanQin Cai, Jian-Feng Cai, and Ke~Wei.
\newblock Accelerated alternating projections for robust principal component
  analysis.
\newblock {\em The Journal of Machine Learning Research}, 20(1):685--717, 2019.

\bibitem{yi2016fast}
Xinyang Yi, Dohyung Park, Yudong Chen, and Constantine Caramanis.
\newblock Fast algorithms for robust {PCA} via gradient descent.
\newblock In {\em Advances in neural information processing systems}, pages
  4152--4160, 2016.

\bibitem{tong2020accelerating}
Tian Tong, Cong Ma, and Yuejie Chi.
\newblock Accelerating ill-conditioned low-rank matrix estimation via scaled
  gradient descent.
\newblock {\em arXiv preprint arXiv:2005.08898}, 2020.

\bibitem{mahoney2009cur}
Michael~W Mahoney and Petros Drineas.
\newblock {CUR} matrix decompositions for improved data analysis.
\newblock {\em Proceedings of the National Academy of Sciences},
  106(3):697--702, 2009.

\bibitem{HH2020}
Keaton Hamm and Longxiu Huang.
\newblock Perspectives on {CUR} decompositions.
\newblock {\em Applied and Computational Harmonic Analysis}, 48(3):1088--1099,
  2020.

\bibitem{chiu2013sublinear}
Jiawei Chiu and Laurent Demanet.
\newblock Sublinear randomized algorithms for skeleton decompositions.
\newblock {\em SIAM Journal on Matrix Analysis and Applications},
  34(3):1361--1383, 2013.

\bibitem{HH2020S}
Keaton Hamm and Longxiu Huang.
\newblock Stability of sampling for {CUR} decompositions.
\newblock {\em Foundations of Data Science}, 2(2):83--99, 2020.

\bibitem{bien2010cur}
Jacob Bien, Ya~Xu, and Michael~W Mahoney.
\newblock Cur from a sparse optimization viewpoint.
\newblock In {\em Advances in Neural Information Processing Systems}, pages
  217--225, 2010.

\bibitem{TroppNystrom}
Joel~A Tropp, Alp Yurtsever, Madeleine Udell, and Volkan Cevher.
\newblock Fixed-rank approximation of a positive-semidefinite matrix from
  streaming data.
\newblock In {\em Advances in Neural Information Processing Systems}, pages
  1225--1234, 2017.

\bibitem{BeckerNystrom}
Farhad Pourkamali-Anaraki and Stephen Becker.
\newblock Improved fixed-rank {N}ystr{\"o}m approximation via {QR}
  decomposition: Practical and theoretical aspects.
\newblock {\em Neurocomputing}, 2019.

\bibitem{HH2019}
Keaton Hamm and Longxiu Huang.
\newblock Perturbations of {CUR} decompositions.
\newblock {\em arXiv preprint arXiv:1908.08101}, 2019.

\bibitem{aldroubi2019cur}
Akram Aldroubi, Keaton Hamm, Ahmet~Bugra Koku, and Ali Sekmen.
\newblock {CUR} decompositions, similarity matrices, and subspace clustering.
\newblock {\em Frontiers in Applied Mathematics and Statistics}, 4:65, 2019.

\end{thebibliography}

\newpage

\appendix

\begin{center}
{\LARGE IRCUR: Supplemental Materials}
\end{center}
\vspace{0.15in}

\section{Setup Details for Numerical Experiments}

We use Matlab R2020a as our testing platform, and all the results are obtained on a laptop equipped with Intel i7-8750H and 32GB RAM. The codes for both AccAltProj and GD are downloaded from the authors' websites, and we manually tuned the parameters for their best performance.
In particular, the actual $\mu$, $\alpha$ and $\kappa$ are provided to AccAltProj and GD for guiding the parameter tuning, while we simply set $\zeta_0=2\|\BL\|_\infty$ for \algoname. To balance between robustness and convergence speed, we pick $\gamma=1.25$ for GD\footnote{Note that GD also uses a parameter named $\gamma$ but it controls a very different operator.}, and $\gamma=0.65$ for AccAltProj and \algoname.  
All algorithms halt when $e_k\leq 10^{-5}$ is satisfied for fair comparison in all the tests.

\subsection{Setup for Synthetic Datasets}

We form the underlying rank $r$ matrix $\BL=\bm{A}\bm{B}^T\in\R^{n\times n}$ via two random Gaussian matrices $\bm{A},\bm{B}\in\R^{n\times r}$. The underlying outlier matrix $\BS$ has uniformly sampled support, and the values of its non-zero entries are i.i.d. uniformly distributed over $[-\mathbb{E}|[\BL]_{i,j}|, \mathbb{E}|[\BL]_{i,j}|]$. Since the data matrices are square, we sample equal number of rows and columns in the following experiments, i.e., $|\cI|=|\cJ|=c_I r\log(n)$, but $\cI$ may not equal to $\cJ$.

\subsection{Setup for Video Background Subtraction}

The two benchmark videos, \textit{shoppingmall} and  \textit{restaurant}, were downloaded from   
\begin{center}
\url{http://perception.i2r.a-star.edu.sg/bk_model/bk_index.html},
\end{center}
and also available at
\begin{center}
\url{https://www.math.ucla.edu/~hqcai/dataset}.
\end{center}
We vectorize and stack the frames of a video to form the data matrix $\BD$ (each column of $\BD$ represents a vectorized frame from the video), then apply RPCA to separate the background (i.e., the low rank component $\BL$) and the foreground (i.e., the sparse component $\BS$). 
For a stable static background, we set $r=2$ and $c_I=4$ in this test. The paramaters of AccAltProj and GD are chosen according to their papers.

\newpage
\section{More Visual Results for Video Background Subtraction Experiment} \label{apx:more experiment result}

\begin{figure}[h]
\centering
\subfloat{\includegraphics[width=.19\linewidth]{1_mall.jpg}} \hfill
\subfloat{\includegraphics[width=.19\linewidth]{IRCUR_F_1_mall_front.jpg}} \hfill
\subfloat{\includegraphics[width=.19\linewidth]{IRCUR_F_1_mall_back.jpg}} \hfill
\subfloat{\includegraphics[width=.19\linewidth]{IRCUR_R_1_mall_front.jpg}} \hfill
\subfloat{\includegraphics[width=.19\linewidth]{IRCUR_R_1_mall_back.jpg}}
\vspace{-0.1in}

\subfloat{\includegraphics[width=.19\linewidth]{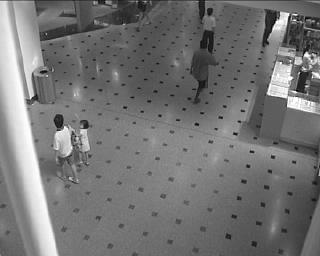}} \hfill
\subfloat{\includegraphics[width=.19\linewidth]{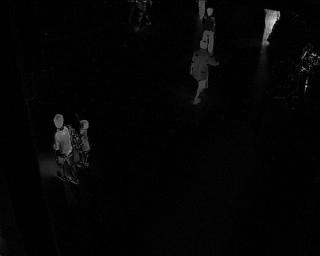}} \hfill
\subfloat{\includegraphics[width=.19\linewidth]{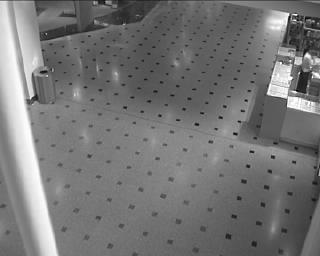}} \hfill
\subfloat{\includegraphics[width=.19\linewidth]{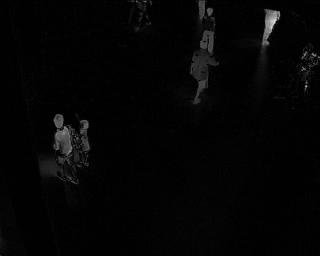}} \hfill
\subfloat{\includegraphics[width=.19\linewidth]{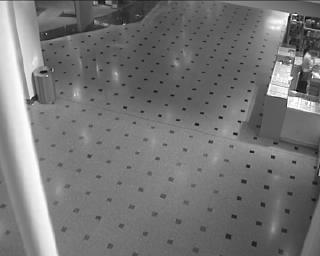}}
\vspace{-0.1in}

\subfloat{\includegraphics[width=.19\linewidth]{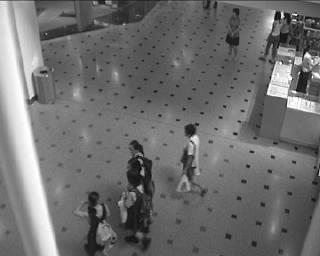}} \hfill
\subfloat{\includegraphics[width=.19\linewidth]{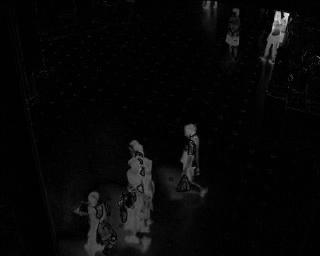}} \hfill
\subfloat{\includegraphics[width=.19\linewidth]{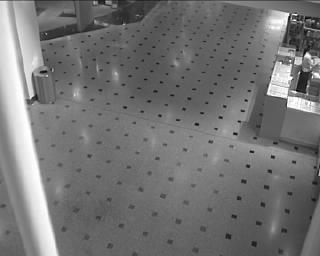}} \hfill
\subfloat{\includegraphics[width=.19\linewidth]{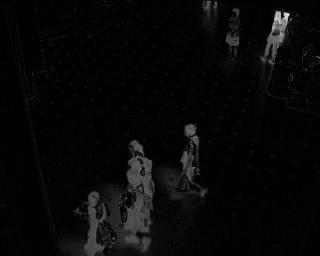}} \hfill
\subfloat{\includegraphics[width=.19\linewidth]{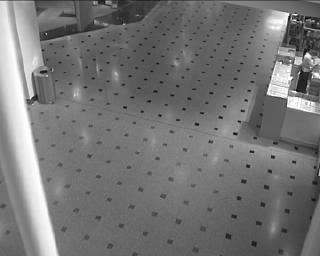}}

\caption{Video background subtraction on \textit{shoppingmall}. Each row is corresponding to a frame. The first column is the original frames. The next two columns are the foreground and background outputted by \algoname-F. The last two columns are the foreground and background outputted by \algoname-R.}\label{FIG:shoppingmall}
\end{figure}

\begin{figure}[h]
\centering
\subfloat{\includegraphics[width=.19\linewidth]{1_rest.jpg}} \hfill
\subfloat{\includegraphics[width=.19\linewidth]{IRCUR_F_1_rest_front.jpg}} \hfill
\subfloat{\includegraphics[width=.19\linewidth]{IRCUR_F_1_rest_back.jpg}} \hfill
\subfloat{\includegraphics[width=.19\linewidth]{IRCUR_R_1_rest_front.jpg}} \hfill
\subfloat{\includegraphics[width=.19\linewidth]{IRCUR_R_1_rest_back.jpg}}
\vspace{-0.1in}

\subfloat{\includegraphics[width=.19\linewidth]{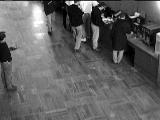}} \hfill
\subfloat{\includegraphics[width=.19\linewidth]{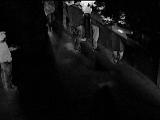}} \hfill
\subfloat{\includegraphics[width=.19\linewidth]{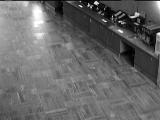}} \hfill
\subfloat{\includegraphics[width=.19\linewidth]{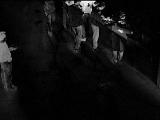}} \hfill
\subfloat{\includegraphics[width=.19\linewidth]{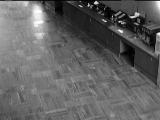}}
\vspace{-0.1in}

\subfloat{\includegraphics[width=.19\linewidth]{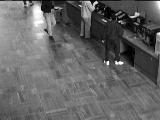}} \hfill
\subfloat{\includegraphics[width=.19\linewidth]{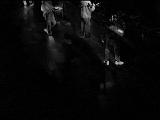}} \hfill
\subfloat{\includegraphics[width=.19\linewidth]{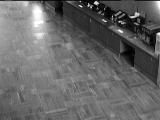}} \hfill
\subfloat{\includegraphics[width=.19\linewidth]{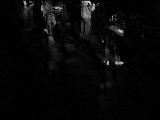}} \hfill
\subfloat{\includegraphics[width=.19\linewidth]{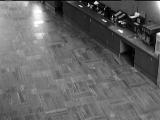}}
\caption{Video background subtraction on \textit{restaurant}. Each row is corresponding to a frame. The first column is the original frames. The next two columns are the foreground and background outputted by \algoname-F. The last two columns are the foreground and background outputted by \algoname-R.}\label{FIG:restaurant}
\end{figure}

As promised, we present more video background subtraction results for \textit{shoppingmall} and \textit{restaurant} in Figures~\ref{FIG:shoppingmall} and \ref{FIG:restaurant}, respectively.

\end{document}